\documentclass[a4paper,twoside]{article}

\usepackage{epsfig}
\usepackage{subcaption}
\usepackage{calc}
\usepackage{amssymb}
\usepackage{amstext}
\usepackage{amsmath}
\usepackage{amsthm}
\usepackage{multicol}
\usepackage{pslatex}
\usepackage{apalike}
\usepackage{tabularx}
\usepackage{SCITEPRESS}     

\begin{document}

\title{Understand Watchdogs: Discover How Game Bot Get Discovered}

\author{\authorname{Eunji Park, Kyung Ho Park and Huy Kang Kim}
\affiliation{School of Cybersecurity, Korea University, Seoul, Republic of Korea}
\email{\{epark911, kyungho96, cenda\}@korea.ac.kr} }


\abstract{The game industry has long been troubled by malicious activities utilizing game bots. 
The game bots disturb other game players and destroy the environmental system of the games. 
For these reasons, the game industry put their best efforts to detect the game bots among players' characters using the learning-based detections. 
However, one problem with the detection methodologies is that they do not provide rational explanations about their decisions. 
To resolve this problem, in this work, we investigate the explainabilities of the game bot detection. 
We develop the XAI model using a dataset from the Korean MMORPG, AION, which includes game logs of human players and game bots. 
More than one classification model has been applied to the dataset to be analyzed by applying interpretable models. 
This provides us explanations about the game bots' behavior, and the truthfulness of the explanations has been evaluated.
Besides, interpretability contributes to minimizing false detection, which imposes unfair restrictions on human players.}

\onecolumn \maketitle \normalsize \setcounter{footnote}{0} \vfill

\section{\uppercase{Introduction}} \label{sec:introduction}
\vspace{-0.1cm}
\noindent Along with the growth of the online game industry, Massively Multiplayer Online Role-Playing Game (MMORPG) has become a significant modern leisure measure.
MMORPG is one of the types of online games that a user creates his or her character to encounter a variety of entertaining content, such as building a social community, selling in-game assets, or even dating with other players. 
As users can experience a wide range of contents in a virtual world, MMORPG ecosystems resemble an actual ecosystem of the real world \cite{galarneau2005spontaneous}.

Interestingly, one of the similarities between the MMORPG ecosystem and the real world is malicious activity. Similar to criminals or offenders in a society, there have existed malicious characters in MMORPG called game bots \cite{castronova2001virtual}.
A game bot is an automated program that makes the character move around the virtual world as specified and performs predefined actions to collect game assets \cite{kang2013online}. 
Because game assets can be traded with the cash in the real world, which is called a Real Money Trade (RMT), most game bots do a repetitive action for asset accumulation, such as hunting weak monsters over and over again without getting tired \cite{kwon2013surgical}. 
By utilizing these advantages of the game bot, some malicious entities form a Gold Farmer Group (GFG), which systematically manages a huge amount of bot characters to earn illegal income on a large scale \cite{kwon2013surgical}.

These malicious activities of GFGs cause damage to the game company. 
First, the amount of game assets offered by GFG is lower than the legal `in-app purchase' price; thus, users do business with GFG.
Furthermore, GFGs harm user satisfaction as they monopolize game items through a massive amount of game bots, and it promotes normal users to leave the game. 
Due to the damage coming from GFGs, the game industry has focused on an effective game bot detection model.

Early approaches employed data mining methods to identify unique characteristics of game bots. 
As game bots are designed to accumulate game assets rather than other activities, bot characters show a different pattern from normal users. 
Past approaches analyzed chatting patterns, social events, or chatting patterns to scrutinize bot characters' distinct patterns. 
Under the effective representation of the features mentioned above, prior researches achieved a significant bot detection performance. 

Although bot detection models with data mining approaches showed a precise performance; however, the performance of the detection model highly relied on the feature engineering process. 
If the bot changes its behavioral pattern in order to evade the monitoring system, a particular feature for the bot detection changes, and it recesses the model performance. 
Not only the change of bots' pattern, but data mining-based models also require frequent improvement along with the update of the game ecosystem. 
In case of the update of the game ecosystem, features used in the detection model might be changed or dismissed; therefore, the practitioner had to provide a huge effort on the feature engineering process.

To overcome the limits, recent approaches present bot detection methods with machine learning models.
Machine learning models effectively recognize patterns with fewer features; thus, prior works employed a wide range of models to the game bot detection \cite{chung2013game,lee2016you}.
Furthermore, deep neural networks contributed to state-of-the-art bot detection performance by recognizing distinct patterns between game bots and normal characters. However, there still exists the problem of interpretability. Although the model identified bot characters from normal users, game operators should provide why a particular character is classified as a game bot. 
Furthermore, interpretable bot detection results produce significant insights that game operators to design the game ecosystem with decreased malicious activities.
Unfortunately, deep machine learning models cannot explain detailed explanations on the detection result and leave a room for improvement related to the interpretability.

This study employed the Explainable Artificial Intelligence (XAI) approaches to establish `explanations' on the game bot detection. 
The XAI is a modern approach to provide detailed logic behind the machine learning models' detection results. 
As a baseline of our study, we established concrete bot detection models that achieve detection accuracy above 88\%. 
We trained two classifiers composed of random forest models (RF) and multi-layered perceptrons (MLP), which are the algorithms that belong to machine learning and deep learning, respectively.
Leveraging trained classifiers, we explored the importance of utilized features on detection. We scrutinized the reasons behind the classification result of two classifiers to provide the interpretability of the model.

For the first classifier with the RF model, we extracted the information regarding feature importance, which is intrinsically equipped in the model. Additionally, we identified permutation importance, which applies feature importance repeatedly. For the second classifier with MLP, we applied Local Interpretable Model-agnostic Explanations (LIME) and SHapley Additive Explanations (SHAP) to scrutinize the explanations about features that stand out most in classifying game bot characters.
Finally, we compared the explanations of two classifiers applied with XAI approaches and addressed how high-performing classifiers identify bot characters from normal characters.

Throughout the research, our contributions are as below.
\begin{itemize}
    \item We established two high-performing bot detection models with RF model and deep neural networks and applied XAI approaches to produce interpretability behind their decisions. Specifically, we figured out significant features on game bot detection leveraging four XAI methods: the feature importance and permutation importance at RF model, and LIME and SHAP at MLP model. Consequentially, we examined both the similarity and difference of significant features between two classifiers.
    
    \item We explored the impact of significant features resulted from XAI approaches to detection accuracy. To estimate the importance of particular features, we deleted a specific feature from the feature set and observed the detection accuracy change. We compared the performance between the feature set without a particular feature and the original feature set and estimated the importance of an excluded feature in the meaning of detection accuracy.
    
    \item We analyzed a difference between game bots and heavy users, which past bot detection studies endeavored to clarify. As heavy users and automated bot characters show a similar pattern, it has been critical to proving the difference between the two types of characters mentioned above. Throughout the analysis with XAI approaches, we checked the difference of significant features between bot characters and heavy users and resulted in candidate features that can be used for further detection analysis.

\end{itemize}

\begin{table*}
\centering
\caption{Related researches of the game bot detection}
\begin{tabular}{|c|c|c|c|}
\hline
Reference & Data & Proposed Method\\
\hline
\cite{kang2013online} & AION & Rule set\\
\cite{chung2013game} & Yulgang Online & Multiple SVM\\
\cite{lee2016you} & Lineage, AION, Blade \& Soul & Logistic Regression\\
\cite{park2019show} & AION & LSTM\\
\cite{tsaur2019deep} & KANO & MLP NN\\
\cite{tao2018nguard} & NetEase MMORPGs & ABLSTM\\
\cite{lee2018no} & Lineage & Network Analysis \\
\hline
\end{tabular}
\label{tab:related work}
\end{table*}

\vspace{-0.1cm}
\section{\uppercase{Literature Review}} \label{sec:literature review}
\vspace{-0.1cm}
\noindent Lots of research have suggested insights for the game bot detection.
In the early stage of the related research, people considered the game bots' characteristics a primary key.
Kang et al. analyzed the bots' behaviors and established detection rules. 
Comparing the game logs of both human players and bots, the researchers discovered that the game bots make a party of two members and stay much longer than average.
They also observed the game bots tend to have lots of experience points logs and fewer race points logs. 
The researchers established bot detection rules and showed efficacy with 95.92\% accuracy based on the knowledge \cite{kang2013online}.
Furthermore, Lee et al. unveiled a monetary relationship between game bots leveraging network analysis. 
By visualizing the bot characters' network in a live service game, they proposed the analysis of transactions among game characters can be a significant pattern of game bot detection \cite{lee2018no}.
Chung et al. proposed a machine learning methodology to detect malicious activities. 
First, twelve behavioral features were selected, and five advanced features were extracted by combining and calculating such features.
The players' data clustered based on game-playing styles: Killers, Achievers, Explorers, and Socializers.
At last, the researchers trained support vector machine (SVM) models with the data clusters to obtain the best model. 
Their methodology demonstrated higher accuracy in all play style clusters with diverse feature combinations compared to the global SVM model \cite{chung2013game}.

Lee et al. built a framework using the self-similarity of players.
The concept of self-similarity came from the fact that game bots have a high chance of repeating the same actions over time.
They preprocessed three MMORPGs dataset to get the self-similarity with other known features. 
The result with logistic regression demonstrated the effectiveness of the feature showing 95\% detection accuracy \cite{lee2016you}. 
However, the methods mentioned above are not sustainable because game bots' actions vary along with their purpose. 
For this reason, if we do not give attention to the feature importance, the machine learning models produce relatively less precise results. 
It makes the feature selection the most crucial step for game bot detection using machine learning models.
Moreover, due to the enormous number of logs by the nature of games, choosing features could be either essential or ineffective with much work. 
In short, the methodologies are excessively dependent on feature selections that require high costs. 

Because of these shortcomings, research has introduced deep learning technologies for bot detection.
Park et al. used the time-series financial data of the game logs.
The insisted the game bots' financial status was exceptional regardless of how the bots behave.
Based on the idea, they used a Long Short-Term Memory (LSTM) model to display the difference between the financial status of human players and bots. 
The LSTM operated well with the time-series data in terms of computations of influences to the future situation. 
They proved the proposed method achieved high performance with over 95\% accuracy \cite{park2019show}.
Tsaur et al. experimented with calculating the abnormal rate of each player and classified game bots through their proposed deep learning models. 
They provided insight into the concept of gray zone players, which are the human players that behave similarly to bots.
Their performance was proven through a real game dataset of ``KANO" \cite{tsaur2019deep}.

Tao et al. proposed a bot detection framework, NGUARD.
The framework included preprocessing, online inference, the auto-iteration mechanism, and offline training.
One of the two solutions provided in the online inference phase was a supervised classification using a pre-trained ABLSTM model.
The model got trained from the offline training phase to classify the game bots well.
The training was conducted based on the continuous sequence data of user behaviors. 
NGUARD framework performed significantly on the bot detection showing higher than 97\% precision \cite{tao2018nguard}. 
Even though both the research of Park and Tao considered several features, they did not require critical feature engineering. 

Table \ref{tab:related work} shows the data and proposed method to detect bots in the previous research.
Bot detection technologies have reached their goal in the game industry, classifying the game bots among human players. 
One problem that has arisen recently is that they cannot explain the classification, which is a chronic problem of deep learning models. 
Because the primary purpose of the research has been a high detection performance, how the models made results has not been in consideration. 
Nowadays, such black-box nature becomes ``hot potatoes" in detection systems. 
In the game bot detection, the most critical feature is not revealed so far.
Through several methods designed to explain non-interpretable models, we compare their explanations to get the best applicable features. 
The following sections describe further details of how to gain interpretability and what affects the bot detection most.

\begin{table*}
\centering
\caption{Features}
\label{tab:features}
\begin{tabularx}{\textwidth}{|c|X|}\hline
Category & Features \\ \hline
Player Information &
  Login count (1), Logout count (2), Playtime (3-4), Money (5), Total login count (6), IP count (7), Max level (8) \\ \hline
Player Action &
  Collect (9), Sit (10-12), Experience (13-15), Item (16-18), Money (19-21), Abyss (22-24), Exp repair (25-26), Portal usage (27-28), Killed (29-32), Teleport (33-34), Reborn (35-36) \\ \hline
Social Activities &
  Party time (37), Guild action (38), Guild join (39), Social diversity (40) \\ \hline
\end{tabularx}
\end{table*}

\section{\uppercase{Experiment and Result}} \label{sec:experiment and result}

\subsection{Dataset}
\noindent We use the game dataset of a popular MMORPG AION provided by the Korean game company, NCSoft.
The company accumulated game log data from AION to discover patterns of malicious players.
The preprocessed data of the given dataset is distributed by Hacking and Countermeasure Research Lab, Korea University. 
The data contains 49,530 players who have played the game for more than three hours, including 6,213 bots.
The data collection period was from April 10 to July 5, with 88 days.

The publisher confirms the ground truth with the data labels, ``Human" and ``Bot."
NCSoft investigated all doubtful players by having people look through the players' game logs manually.
Even after classifying humans and bots in such a method, the company kept updating labels for players who got confirmed not to be bots through customer inquiries or other ways.
One more thing to consider is that the numbers of data in the two classes differ enough to make the classifiers biased. 
To eliminate the bias, we proceed with both over-sampling and under-sampling by manipulating the number of each class' data.

\subsection{Features}
\noindent The raw dataset included an enormous amount of game log data that expressed every player's characteristics related to the items, skills, and social activities in a time-series manner.
We decided to use a preprocessed dataset that converted time series data into table structure data instead of the original dataset because we only need information about each account and its characteristics for our research purposes.
Additionally, the data of players with level below five were disregarded because they have not experienced the game enough.

The list of features of the dataset is described in Table \ref{tab:features}.
The player information category contains information related to the users' gameplay, and the player action category covers features on the users' behaviors while playing the game.
The category for social activity shows how much the users have participated in social activities such as party and guild to interact with other users in the game.

\begin{figure*}[t]
\centering
\begin{tabular} {c c}\\
\includegraphics[width=0.47\textwidth]{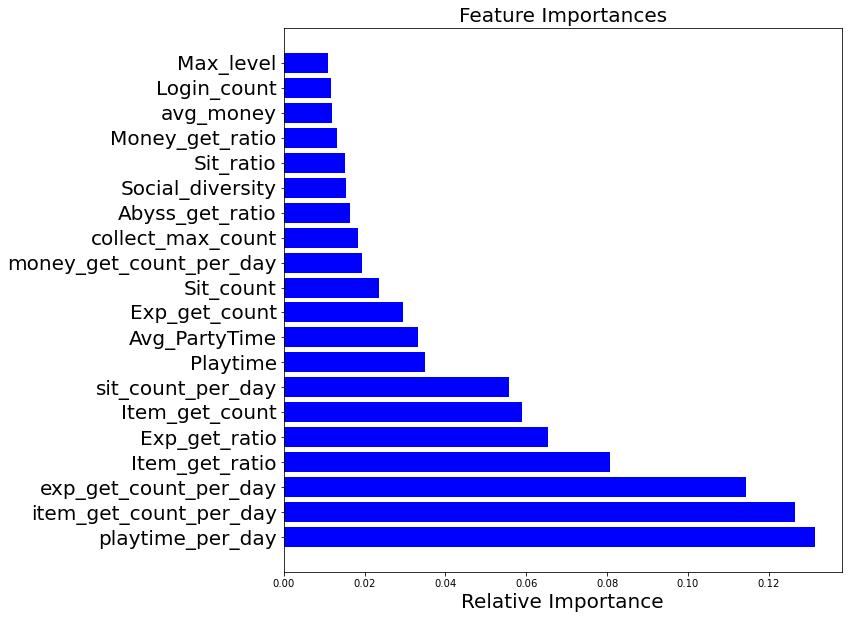} & 
\includegraphics[width=0.47\textwidth]{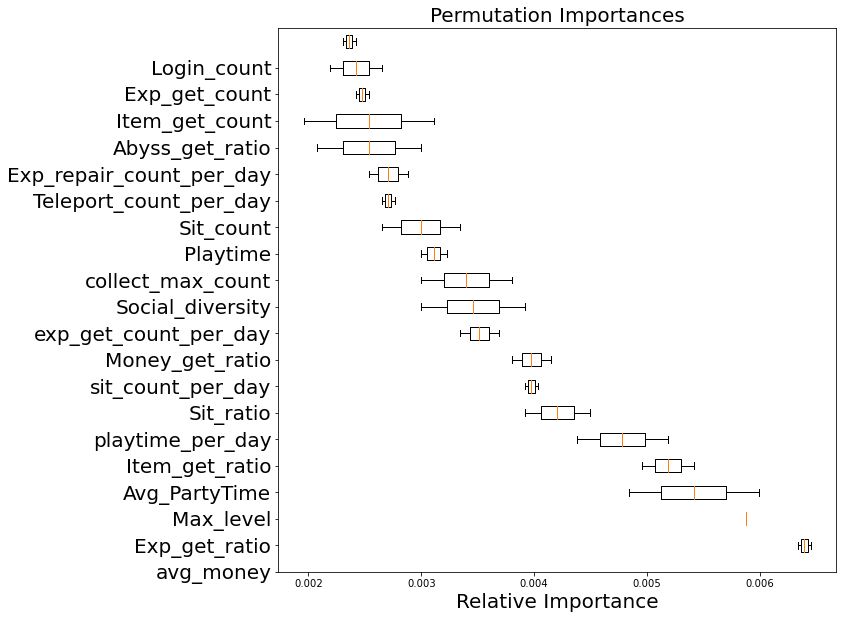} \\
(a) Feature importance & (b) Permutation importance\\
\end{tabular}
\caption{Explanations for Random Forest Classifiers}
\label{fig:Feature Importance}
\end{figure*}

\subsection{Experiment Setup}
\subsubsection{Baseline Classifiers}
\noindent For the detection, we used RF and MLP models.
RF is a tree‐based classifier, and it is one of the traditional machine learning models for classification. 
The model randomly selects combinations of features for the most acceptable performance. 
RF is not in a black box; thus, we choose the RF for the experiment with a baseline assumption that the RF model can provide rational explanations.
On the other hand, MLP is considered as the most common deep learning model. 
Our experiment uses the MLP model because it is not limited to specific data forms, and it gives a chance to try diverse layers to customize the model. 
We organize the model with one input layer, three hidden layers, and one output layer that classifies the player into three classes. 
As this paper does not aim to build an accurate classification model, the model parameter is not specifically selected.

\subsubsection{Explainers}
\noindent The purpose of the experiment is to conduct several explainable models. The RF model can explain itself with ``feature importance" attribution. 
Considering the entire decisions that the model makes, it computes the weight of each feature. 
The model estimates the influence of features based on the impurity.
It explains how the RF model predicted and which features performed essential roles in the classification.
Another explainer derived for the RF classifier is an inspection method called ``permutation\_importance."
The permutation is computed the feature importance through multiple validation and evaluations. 
In the case of the MLP classifier, it guarantees a higher accuracy in the detection process, but the model cannot be interpretable.
For the interpretability of the MLP classifier, LIME is a known open-source module that serves the explanation for the single prediction. 
Ribeiro et al. introduced LIME which gets importances of each feature by changing the input data for the specified instance and getting into the different class \cite{ribeiro2016should}.
Unlike LIME, SHAP supports the explanation of the global instances. 
Making the trained MLP model go through SHAP provides us a perception of the overall data classification.

\subsubsection{Experiments}
\noindent A comparison between definitive explanations for RF Classifier and external explanations using LIME and SHAP modules for MLP is the key takeaway.
First of all, it is essential to train the models with RF and MLP to classify the human players and game bots with decent performance.
We build and train a simple MLP model with a customized number of hidden layers and demonstrate acceptable performance. 
When the classifiers are trained enough, it is time to explore which features influence the decision making on classifications.
The RF model uses its built-in attribute, the feature importance, and is applied to calculate the permutation importance of the features. 
The outputs give us what features impact on the classifier most. 

On the other hand, as the black-box model, the model with MLP requires interpretability coming from the external module named LIME. 
The primary LIME explains for one instance at a time, and it guarantees the accurate explanations.
However, observing the interpretability for all instances is unrealistic in the industrial environment; thus, a LIME extension called an SP-LIME is used.
SP-LIME selects representative cases with diverse explanations.

After the experiment, we obtain a total of four explanatory results and attempt to evaluate them.
We delete the most significant features shown in each result and retrain the classifiers to compare the efficacy of the classification.
The more accurate the XAI explains, the more performance we lose when eliminating the features.

Furthermore, the last experiment extracts only false positives from the classification.
The false-positive case is that the model classifies human users as a bot, and for the game industries, the case should be minimized. 
We investigate the cases through LIME and figure out the characteristics of human players who play like bots, ``Heavy users.''
Based on significant features on misclassifying heavy users to bots, the class named `Human' is divided into normal users and heavy users.
The classification problem now has three categories, and we expect to get lower false positives and a better detection rate.

\subsection{Results}

\begin{table*}
\centering
\caption{SP-LIME Results}
\begin{tabular}{|c|c|c|c|}
\hline
Class & Feature 1 & Feature 2 & Feature 3\\
\hline
Human 1 & Login\_day\_count & Item\_get\_ratio & Login\_count\\
Human 2 & Login\_count & Exp\_get\_count & playtime\_per\_day\\
Bot 1 & Exp\_get\_count & Login\_day\_count & exp\_get\_count\_per\_day\\
Human 3 & Item\_get\_ratio & Teleport\_count\_per\_day & Login\_day\_count\\
Bot 2 & exp\_get\_count\_per\_day & sit\_count & Login\_day\_count\\
\hline
\end{tabular}
\label{tab:sp-lime}
\end{table*}

\noindent At first, we trained a RF classifier and MLP classifier. 
The dataset contained two classes: Bot and Human, and it has been through the oversampling process due to the low number of data in the ``Bot" class.
90\% of the dataset was used to train the models, and the rest of the data was for the validation.
The validation accuracy of the RF model is 88.51\%, while the MLP model reached over 90.10\%.

\begin{figure}[]
\centering
\begin{tabular} {c}\\
\includegraphics[width=0.45\textwidth]{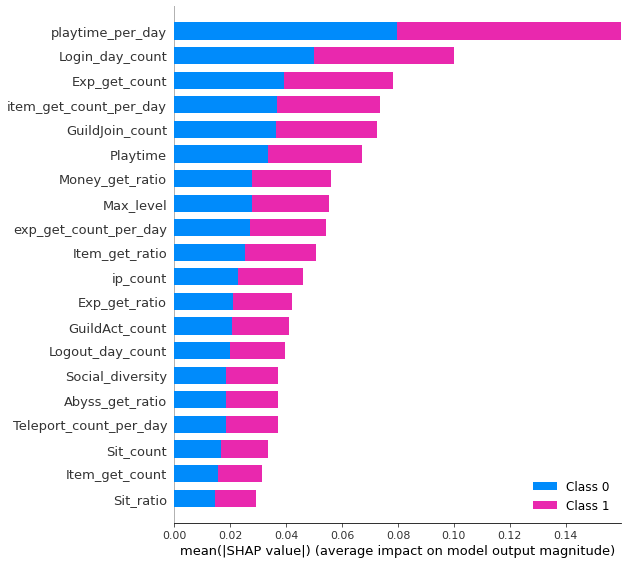}\\
\\
\end{tabular}
\caption{Explanations using SHAP for MLP}
\label{fig:SHAP}
\end{figure}

The first explanation method is `feature\_importances\_,' which is for the RF model.
Figure \ref{fig:Feature Importance} shows the result of the explanation method with the top 30 crucial features on its left side. 
According to the first method, the most significant feature of the RF model is `playtime\_per\_day.'
`item\_get\_count\_per\_day' and `exp\_get\_count\_per\_day' follow in it, and features about item and playtime are on the top of the rank.
The other explanation method at the RF model is `permutation\_importance.'
The importance of playtime\_per\_day has shown much higher in this method, and we thought a reason is that this method repeats the calculation of representative features to obtain the permutation.
Therefore, the feature showed extraordinary importance; thus, we exclude the feature on the right side of Figure \ref{fig:Feature Importance}, showing 30 importances from the second important feature.
Unlike the previous method, the money-related features and points getting ratio seemed to be influential.

On the other hand, the deep learning-based classification model does not provide a method to describe the model independently.
To detour the disadvantage, we applied the LIME module on the model.
Because LIME provides interpretability of what classification results in each case, it is not practical to scrutinize all the cases one by one.
Therefore, an extended version of LIME called SP-LIME has been introduced to analyze several representations, and we chose to examine the five cases.
The results are in Table \ref{tab:sp-lime}, saying the Human 1 frequently logs in a day and is mainly focused on collecting items. Human 2 involves the players who have played the game for a long time. The Human 3 appears to be wandering around the game world, and it has a similar play style to the Human 1. On the other hand, the common characteristic of Bot classes is that they gain an overwhelming amount of experience points in a day compared to human players.
After all, SHAP looks for explanations by checking each instance like LIME, and it also provides an average global explanation.
Applying SHAP to the trained MLP classifier, Figure \ref{fig:SHAP} shows its output with the crucial features.
It represents the importance of the classification features without the distinction between positive and negative values \cite{lundberg2017unified}.
The result said the playtime had the most influence on the classification, and login counts and experience points getting counts also had notable impacts.

At the results, playtime\_per\_day and item\_get\_count\_per\_day are the most important in the feature importance method, and avg\_money and exp\_get\_ratio are the most important in permutation importance.
The most frequently observed features in SP-LIME results were login\_day\_count and exp\_get\_count, while SHAP said playtime\_per\_day and login\_day\_count were critical.
The two classification models got trained again by excluding the most important features from the corresponding XAI results.
Removing playtime\_per\_day and item\_get\_count\_per\_day features, the performance of the RF classifier has not reduced at all with an accuracy of 88.58\%.
With the permutation importance, when we deleted avg\_money and exp\_get\_ratio from the data, the classifier showed a little lower performance, 88.21\%. 
For the MLP classifier, we first dropped login\_day\_count and exp\_get\_count according to SP-LIME results, and its performance with validation accuracy became 89.13\%, which were lower compared to the original performance of 90.10\%.
In the end, when deleting playtime\_per\_day and login\_day\_count features referring SHAP, the MLP classifier showed 89.37\% validation accuracy. 

\begin{figure}[]
\centering
\begin{tabular} {c}\\
\includegraphics[width=0.4\textwidth]{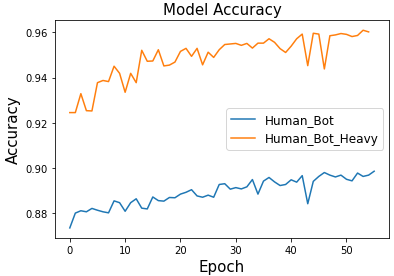}\\
\\
\end{tabular}
\caption{Comparison of classification performance by class distribution}
\label{fig:heavy}
\end{figure}

The last experiment was to identify features that play an important role in distinguishing between the heat users who play like bots.
Using only the MLP classifier, when the trained model got obtained, we extracted only false-positive cases from the classification results, and the features that played important roles in the cases were identified through XAI.
Because we used an MLP classifier and required XAI methods that can represent explanations for each instance, we decided to use LIME for this purpose. 
We examined several cases that the classifier predicted as a bot when the case was a human user. 
In the most cases, exp\_get\_ratio feature seemed effective, followed by collect\_max\_count and exp\_get\_count\_per\_day. 
Based on the results, the new class called ``Heavy user" was formed from the ``Human'' category, and the classification used the new dataset with three classes. 
The classification performance of the MLP model improved a lot from the original MLP with the validation accuracy of 96.02\%.
Furthermore, the metric of false positives became significantly lower from 881 to 327. 
Figure \ref{fig:heavy} describes the improvement by classifying heavy users utilizing the XAI.

\vspace{-0.1cm}
\section{\uppercase{Discussion}} \label{sec:discussion}
\vspace{-0.1cm}
\noindent We have identified the two base classification models' performance and features that played an essential role in the classification through the experiments.
The interpretability of the RF model could only identify the rank of feature importance, but not the difference between the game bots \& the non-bot.
Consequently, we used LIME and SHAP, and they successfully supported the functionalities. 

First, we could identify the features that were important in classifying bots.
The RF classifier explained the binary result that is either human or bot, while the MLP model provided an explanation based on the prediction probabilities of each class.
We found out that the main criteria for distinguishing bots are related to earning items and experience points through explanatory technologies.
Additionally, playtime\_per\_day showed significant relevance in some explanations.

The second experiment was purposed to determine which method is the best among the XAI techniques by observing the reduction of the performance when deleting the most significant features. 
The explanation from feature importance could not make the classification performance lower, while one from the permutation importance affected the performance with a little figure, 0.3\%. 
We considered the permutation importance showed a better explanation than the feature importance because of its repetitive patterns.
The important features of the MLP model seemed to be more apparent than the former model. 
When we deleted the features that SP-LIME determined as the most important, the detection accuracy of the classifier became 1\% lower. 
With the result of SHAP, the performance change was similar to the experimental result using permutation importance. 
Therefore, we could observe some tendencies with the essential features of bot detection.
Even though playtime\_per\_day appeared a lot in the explanations, it has little effect on the bot detection itself. 
The features related to the day\_count or get\_ratio of login, experience points, items are rather more critical. 
Additionally, among the four explainable AI techniques, the one that provided the best explanations was the LIME module. 

At last, we made a distinction between regular users and heavy users. 
With the three features, exp\_get\_ratio feature, collect\_max\_count and exp\_get\_count\_per\_day, ``heavy user'' class was created, and the classification showed outstanding performance with low false positives.
The result implies that the features were essential to distinguish the heavy players and game bots. 
We added the heavy user class by identifying the classification reasons and confirmed that the classification performance had improved a lot without model modification.

Whether it is a game bot or a human player has been a long time problem.
Additionally, false positives can cause complications if a human player is misclassified as a bot.
These days, with more deep learning-based classifications, such characteristics and problems have become more noticeable.
There is a growing need to discover the cause of detection for those who want to protest false detections as bots.
To that end, we checked out how to explain the bot detection models.
In this way, the game industries can verify the properties of the game bot and examine the features that must be referred to.
By investigating how the game bot detection works, designing a new game will help identify the data extracted to win the fight against the game bot.
We also think it is useful when the game logging system gets updated with better features.

\section{\uppercase{Conclusion}} \label{sec:conclusion}
\noindent Numerous researches have suggested methods for detection to resolve the problem that the online game industry is suffering from.
However, because a game industry is based on the interactive behaviors of real people, false detections should not be overlooked when it occurs. 
The actions to provide logical evidence are the most required for this reason. 
There has been a disadvantage of not interpretable in a complicated deep neural network model, but it has become feasible through a new paradigm called XAI.

Upon the importance of interpretability, we aimed to establish a bot detection model that pursues two goals: a precise detection performance and providing cues of detecting game bot activities. 
We extracted candidate features from raw log data of AION, a popular MMORPG in a live service. 
Based on extracted features, we applied multiple XAI approaches to the bot detection task. 
First, we leveraged a RF model, a classical machine learning algorithm, to provide the importance of particular features and permutations.
Then we set a simple MLP model with XAI modules: LIME and SHAP.

Along with the experiment, a game bot detection performance of both the RF model and MLP model achieved over 88\% accuracy, which is a detection result as a baseline. 
We compared the resulted set of features from both the RF Model and MLP model with XAI modules. 
We also evaluated the XAI methods by comparing the detection performance after excluding significant features from the feature set.
The permutation importance and SHAP modules were evaluated as better methods than the feature importance, and the LIME module was rated as the best of the used explanation methods. 
Last but not least, we also clarified the difference between game bot characters and heavy users. 
As game bots and heavy users similarly accumulate in-game assets in a short time, past detection models experienced confusion between two classes. 
We established our analogy based on the XAI results, and we evaluated it resolved the confusion through the experiment. 

We believe future studies consider the in-depth examination of the significant features that the explanation models present.
It involves exploring the difference between the meanings of significant features when using the RF model's explanation and the XAI module.
Even the data is from different domains other than the game industry, if we analyze the meaning of each feature, it will reflect our understanding of the domain and reach a way to reduce false detection further.

\vfill


\end{document}